\documentclass[11pt, a4paper]{article}
\usepackage{amsmath}
\usepackage{amssymb}
\usepackage{amsthm}
\usepackage[margin = 1.2in]{geometry}
\usepackage{graphicx}
\usepackage{hyperref}
\usepackage[utf8]{inputenc}
\usepackage{lipsum}
\usepackage{mathrsfs}
\usepackage{parskip}
\usepackage{setspace}
\usepackage{soul}
\usepackage{tikz}
\usepackage{url}
\usepackage{float}
\usepackage{amsmath}
\usepackage{enumerate}
\usepackage{subcaption}
\usepackage[numbers,sort&compress]{natbib}

\newtheorem*{hypothesis}{Lottery Ticket Hypothesis}

\begin{document}

\begin{titlepage}
    \begin{center}
        \vspace*{1cm}
            
        \Huge
        \textbf{Transferability of Winning Lottery Tickets in Neural Network Differential Equation Solvers.}
            
        \vspace{0.5cm}

        \vspace{1.5cm}
        \large
        \textbf{Edward Prideaux-Ghee}\\
        \vspace{1cm}
        \Large
        June 2023\\
            
        \vfill
        \begin{abstract}
        \noindent
        Recent work has shown that renormalisation group theory is a useful framework with which to describe the process of pruning neural networks via iterative magnitude pruning. This report formally describes the link between RG theory and IMP and extends previous results around the Lottery Ticket Hypothesis and Elastic Lottery Hypothesis to Hamiltonian Neural Networks for solving differential equations. We find lottery tickets for two Hamiltonian Neural Networks and demonstrate transferability between the two systems, with accuracy being dependent on integration times. The universality of the two systems is then analysed using tools from an RG perspective.
        \end{abstract} 
        
        \vspace{1cm}
        \large

    \end{center}
\end{titlepage}

\newpage

\newpage
\doublespacing
\tableofcontents
\singlespacing

\newpage

\section{Introduction}
Within the past few decades, machine learning has developed into an ever more popular and successful method in solving a variety of problems in fields such as computer vision and natural language processing (NLP). A subset of these methods is deep neural networks (DNNs) in which potentially billions of parameters are trained to find an optimal solution to a problem \citep{Schmidhuber_2015}. The large size of these networks means they demand extensive computation power as well as a lot of memory. However, many DNNs are over-parameterised \citep{dauphin2013big} which means that there is unnecessary computing power being used. This has led to research into the Lottery Ticket Hypothesis \citep{frankle2019lottery} - a form of transfer learning - which states:  
\begin{hypothesis} A dense neural network contains a sparse subnetwork that, when trained under the same initialisation, can perform with similar or even better accuracy than the original network when trained for the same number of epochs.
\end{hypothesis}
One current method for finding sparse subnetworks with potentially similar performance to the full network is through a process of pruning: removing weights from the network. There are many different types of pruning procedures, such as magnitude based, random and single-shot \citep{vadera2021pruningmethods},\citep{lee2019snip},\citep{yu2021hessianaware},\citep{hayou2021robust},\citep{liu2020pruning},\citep{Wimmer_2023} with very little consensus on which methods should be used in different situations. In this report, we will focus on one common technique - Iterative Magnitude Pruning (IMP) - as it has been shown to give a "state-of-the-art sparsity-accuracy trade-off" \citep{https://doi.org/10.48550/arxiv.1902.09574}.
Winning Lottery Tickets have been found in several areas such as computer vision \citep{frankle2019lottery},\citep{chen2020lottery},\citep{vandersmissen2023layerLTH}\citep{chen2021lotteryCV} and natural language processing (NLP) \citep{michel2019sixteenNLPpruning} however the methods used to find them are very computationally expensive. Therefore, research has been done into the universality of winning lottery tickets across different tasks. The existence of universal lottery tickets has been proven \citep{burkholz2022existence} suggesting that they could be a general occurrence across several fields. In fact, transferability of winning lottery tickets has been shown in several fields \citep{kong2022playing},\citep{mehta2019sparse},\citep{morcos2019ticket} including across networks with different architectures \citep{chen2021elastic}. 
Until recently, there was very little theory around the transferability of lottery tickets and therefore transfer experiments were the only indication as to whether it would be successful or not. However, work by Redman et.al.\citep{redman} has provided a link between sparsifying neural networks and renormalisation group (RG) theory. Specifically, it has been shown that IMP is a renormalisation group scheme and therefore tools from RG theory can be used to study the LTH. RG theory has historically been used predominantly in statistical mechanics, explaining universal behaviour in thermodynamic systems during a phase transition \citep{Goldenfeld92}. One important parallel between RG theory and pruning DNNs is the concept of power-law scaling. During a phase transition, variables scale via a power-law relationship with a critical exponent. The same behaviour has been observed in IMP \citep{https://doi.org/10.48550/arxiv.2006.10621} with error scaling via the following relationship within a critical parameter density 
\begin{equation}
    \label{crit_exp}
    \epsilon \sim cd^{-\gamma}
\end{equation}
where $\epsilon$ is error, $d$ is density and $\gamma$ is a constant. 

One area where the LTH and universality have been less explored is that of scientific solvers. These address a range of problems, such as handling high-dimensionality \citep{Weinan_E_2017_High_dim},\citep{SIRIGNANO20181339DGM},\citep{guidetti2021dnnsolve}\citep{ZENG2022111232}, and have been applied to a wide range of situations \citep{PINNS},\citep{Lu_2021_deep-o-net},\citep{ba2023shaper},\citep{MICHOSKI2020193},\citep{shen2020deep},\citep{Brunton_2016}.
Specifically, we will be looking into neural network differential equation (DE) solvers. These aim to solve an optimisation problem to provide an analytic, closed form differentiable function as a solution \citep{Lagaris_1998_NNDEs}. These types of solvers have been studied individually \citep{dogra2020dynamical} \citep{Mattheakis_2022} \citep{greydanus2019hamiltonian} \citep{BERG201828} but there exists very little knowledge of universality between solvers. As an example, we take Hamiltonian Neural Networks \citep{greydanus2019hamiltonian} as they provide useful test cases which should hopefully give an insight into the wider nature of scientific solvers.

\vspace{1.5cm}
\section{Renormalisation Group and Scaling}

In this section, we aim to formally define a renormalisation group (RG) so that we may later use its related tools for our case of understanding IMP and the LTH.
First, we use the example of block spins in the 2D Ising model from thermodynamics to introduce the concept of power scaling macro-observables near a phase transition \citep{RG_scaling}. This idea is then developed further to give the concept of the renormalisation group and universality. The development of this theory will follow the work of Goldenfeld (Lectures on Phase Transitions and the Renormalization Group, chapter 9, p229-p256) \citep{Goldenfeld92} which views renormalisation group from a thermodynamic phase transition perspective. For a quantum field theory perspective see \citep{renormalisation2}.

Conceptually, the procedure of the renormalisation group as thought of by Kandoff contains three steps. The first step is course-graining which involves reducing the resolution of the system by increasing the smallest length scale from $a$ to $la>a$. \citep{Simons97}. Then in order to restore the original resolution, the system is rescaled by reducing all length scales by factor $l$. The final stage is to renormalise the system so that variables vary on the same scale as originally.

\subsection{Ising Model and Block Spins} 

Consider a system of N spins arranged on a 2-dimensional square lattice with spacing $a$. For now, we make the assumptions of locality and rotational and translational symmetry in the Hamiltonian. That is each spin only interacts with an external magnetic field and its nearest neighbours, with couplings determined by the constants $k_1$ and $k_2$ respectively. In this case, we have the Hamiltonian of the system \citep{Goldenfeld92}:
\begin{equation}
    \mathcal{H}(\textbf{s,k}) = -\sum_i k_1s_1 - \sum_{\langle i,j \rangle}k_2s_is_j
\end{equation}

Where $s_i$ are the spins $s_i \in \{-1,+1\}$, $k_1$ and $k_2$ are the strengths of the coupling constants and $\langle \cdot , \cdot \rangle$ denotes the nearest neighbours in the lattice. This is the simplest model except the trivial case where atoms only interact with themselves. As we are interested in behaviour near critical points and phase transitions for our analysis of IMP, we denote the singular part of the free energy of the system near the phase transition as $f_s(t,h)$\citep{Goldenfeld92}. Within the vicinity of the phase transition, we must consider quantities on the length scale of the correlation length $\xi$, which describes the length scale on which variables are correlated.

The system can then be investigated at different scales by using a process of "course graining" to view a block of multiple spins as one singular spin with properties aggregated from the constituent sub-spins \citep{Goldenfeld92}. If we take a block of spins of side length $al$, we then have a system of $Nl^{-2}$ blocks each containing $l^2$ individual spins. We define this block-spin transformation in the following way:
\begin{equation}
    s' = \frac{1}{|\Bar{m}_l|}\frac{1}{l^2} \sum_{i\in I}s_i
\end{equation}
where I is the set of indices of spins in the new block and $\Bar{m}_l$ is the average magnetisation of the block defined by
\begin{equation}
    \Bar{m}_l = \frac{1}{l_2}\sum_{i\in I} \langle s_i \rangle.
\end{equation}
Re-normalising the block spins like this ensures that they can take the same values as the original individual spins ($\pm1$). Now we assume that, similar to the original system, each block spin interacts only with its nearest neighbouring blocks and the external field. Hence we can define new coupling constants $k_i'$ which determine the nature of these interactions. This assumption leads us to finding the new effective Hamiltonian of the system to be:

\begin{equation}
    \mathcal{H}(\textbf{s',k'}) = -k_1'\sum_i s_i' - k_2'\sum_{\langle i,j \rangle}s_i's_j'
\end{equation}
which is of the same form as the original Hamiltonian. Considering the correlation length $\xi$, initially we had $\xi_1$ measured on a scale of the lattice spacing $a$. After performing the course graining, we now have $\xi_l$ which is measured on a scale of the new spacing between blocks $la$. This therefore gives the relation $\xi=\xi_lla=\xi_1a$ which gives:
\begin{equation}
\label{correlation_length_relation}
\xi_l = \xi_1/l    
\end{equation}
One key property near the critical point during phase transition is the divergence of the correlation length $\xi \rightarrow \infty$. As $l>1$, $\xi_l<\xi_1$ which means the new system is further from the critical point than in the original system, and hence the new system is at an effective reduced temperature $t'$ and magnetisation $h'$ \citep{Goldenfeld92}. 
Combining these results, the form of the free energy of the effective system $f_s(t',h')$ is the same as the original system but with new reduced temperature and magnetisation variables.
\begin{equation}
\label{fs_eq}
    f_s(t',h') = l^{2}f_s(t,h)
\end{equation}
Therefore, the process of course graining has retained the structure of the system, although with scaled parameters. So, denoting this block spin transformation as $\mathcal{R}$, we can describe the effect on the Hamiltonian as:
\begin{equation}
    \mathcal{R}\mathcal{H}(\textbf{s,k}) = \mathcal{H}(\textbf{s}',\mathcal{T}\mathbf{k}) = \mathcal{H}(\textbf{s',k'})
\end{equation}
Where $s',k'$ are the new spins and coupling constants respectively, and the mapping of the coupling constants is denoted by $\mathcal{T}: \mathbb{R}^K \rightarrow \mathbb{R}^K$. As $\mathcal{R}$ is repeatedly applied, we get a flow of coupling constants, known as the RG flow, which is determined by the eigenvectors of $\mathcal{T}$ when linearized near a fixed point. The corresponding eigenvalues are classified into three classes: relevant if $\lambda_i > 1$; irrelevant if $\lambda_i <1$; or marginal if $\lambda_i = 1$.
\\
\\
As our interest is in understanding the power law scaling, we further assume that the temperature and magnetism in the critical region transform under course graining by:
\begin{equation}
\begin{split}
    t' = tl^{y_t} \hspace{.3in} y_t>0\\
    h' = hl^{y_h} \hspace{.3in} y_h>0
\end{split}
\end{equation}
Inputting this into (\ref{fs_eq}) gives us
\begin{equation}
\label{fs_eq2}
    f_s(t,h) = l^{-2}f_s(tl^{y_t},hl^{y_h})
\end{equation}

As there are no restrictions on the level of scaling being implemented ($l$), taking $l$ such that $|t|l^{y_t}=1$ allows for the convenient form:

\begin{equation}
    f_s(t,h) = |t|^{d/y_t}f_s(1,h|t|^{-y_h/y_t})
\end{equation}

This is solely a function of $h|t|^{-y_h/y_t}$ and hence can be written
\begin{equation}
    f_s(t,h) = |t|^{d/y_t}F_f(h|t|^{-y_h/y_t})
\end{equation}
where $F_f(x) = f_s(1,x)$. This can therefore be written in the form of the static scaling hypothesis by denoting $2-\alpha = \frac{d}{y_t}$ and $\Delta = \frac{y_h}{y_t}$:
\begin{equation}
    f_s(t,h) = |t|^{2-\alpha}F_f(h|t|^\Delta)
\end{equation}

This is a specific example of how a course-graining procedure can lead to the observed scaling of macro-observables by discounting the irrelevant couplings between degrees of freedom on a smaller scale. 

\subsection{RG}
The above example of block spins in the 2D Ising model demonstrates how it makes sense to take this form of the scaling for certain variables when near the critical point. However, the assumptions made cannot make sense when extended to different situations. For example, in the case where spins do not interact with their neighbours but do interact with spins further away. In this case, it would not make sense for block spins, comprised of multiple spins, to not interact with their neighbours. Therefore, the concepts of renormalisation group theory are now developed for a more general Hamiltonian.

In order to concretely define a renormalisation group, the course-graining process used above must be formally analysed, and the origin of singular behaviour must be understood.
As in the block spin case, the fundamental concept is that if given a system with interactions on a scale $a$, we take blocks of length $la$, then rescaling and normalising the resulting system gives one similar to the original system with respect to the degrees of freedom. This gives us the characteristic that slightly different systems (e.g interactions between nearest neighbours/next-nearest neighbours) can have the same effective Hamiltonians after course-graining which is important in the context of universality classes. For example, in the block spin case where individual spins do not interact with their nearest neighbours, the resulting system after course-graining could be similar to that where individual spins do. 

Now we consider Hamiltonians of the general form 
\begin{equation}
    \mathcal{H} = \sum_n K_n \Theta_n\{S\}
\end{equation}
with coupling constants $K_n$, and $\Theta_n$ local operators as functions of the degrees of freedom $\{S\}$ \citep{Goldenfeld92}. Now we perform course-graining similarly to before, collecting degrees of freedom together in a linear block of length $la$. We call this general transformation a "renormalisation group transformation" $R_l$. As before, we are free to choose the level of course-graining $l$ so can denote the transformation of coupling constants: 

\begin{equation}
    [\textbf{k}'] = \mathcal{R}_l[\textbf{k}]
\end{equation}

The renormalisation transformations $R_l$ in fact form a semi-group where successive transformations with course-graining scales $l_1,l_2$ are equivalent to a single transformation with scale $l_1 \cdot l_2$ \citep{Goldenfeld92}:
\begin{equation}
    R_{l_1l_2} [K] = R_{l_2} \circ R_{l_1} [K] 
\end{equation}

In order to calculate $R_l$, we now properly define the course graining procedure. First we will need to utilise the partition function $Z_n$ and quantity $g$ relating to free energy per degree of freedom.
\begin{equation}
\begin{split}
    Z_N[K] = & Tr \hspace{.1in}e^\mathcal{H} \\
    g[K] = & \frac{1}{N} \log Z_N[K]
\end{split}
\end{equation}

In order to perform the course graining process and reduce the number of degrees of freedom, we perform a partial trace over the degrees of freedom. 

\begin{equation} \label{RG_eff_H}
\begin{split}
e^{\mathcal{H}'_N\{[K'],S_I'\}} & = Tr'_{\{S_i\}}e^{\mathcal{H}_N\{[K],S_I\}} \\
 & = Tr_{\{S_i\}}\mathcal{P}(S_i, S_I')e^{\mathcal{H}_N\{[K],S_I\}}
\end{split}
\end{equation}
Where $Tr_{\{S_i\}}$ is the trace operator over the values that $S_i$ can take \{$\pm1$\}. Here, to allow the trace to be unrestricted, a projection operator $P(S_i,S'_I)$ is used. This is constructed in such a way that the block degrees of freedom $S_I$ take the same values as the original degrees of freedom, thus maintaining the characteristics of the original system. 

As an example of how the projection operator is constructed, we use the case of the Ising 2D system as previous explained. Having Ising spins on a 2D lattice, we use an RG transformation with blocks of length $(2l+1)a$ so that there is an odd number of spins $(2l+1)^2$ in each block. In order for the block spins have the same values as the original degrees of freedom, we define
\begin{equation}
    S'_I = sign(\sum_{i \in I}S_i) = \pm 1
\end{equation}

Therefore, there is the associated projection operator

\begin{equation}
    P(S_i, S'_I) = \Pi_I \delta ( S'_I - sign(\Sigma_{i \in I} S_i))
\end{equation}

where $\delta$ is a Kronecker delta function. Clearly this is not a unique RG transformation that ensures $S'_I = \pm 1$. However the projection operator must satisfy the following three properties \citep{Goldenfeld92}:

\begin{enumerate}[i]
\label{conditions}
    \item $P(s_i, s_I') \geq 0$
    \item $P(s_i, s_I')$ respects the symmetries of the system
    \item $\sum_{\{s_I'\}} P(s_i, s_I') = 1 $
\end{enumerate}

Condition (i) is necessary to ensure that the exponential of the effective Hamiltonian $e^{\mathcal{H}'_N\{[K'],S_I'\}} \geq 0$ and is thus $\mathcal{H}'$ is well defined as the effective block spin Hamiltonian.   

The second condition (ii) guarantees that there exist no new forms of couplings or symmetries in the new system that were not possible in the original system i.e. the effective Hamiltonian can be written in the same form as originally, but with different values of coefficients. i.e. Given a system of N degrees of freedom described by Hamiltonian:
\begin{equation}
    \mathcal{H}_N = NK_0 + h\sum_i S_i + K_1\sum_{i,j}S_iS_j + ...
\end{equation}
the effective Hamiltonian can be written in an equivalent way:
\begin{equation}
    \mathcal{H'}_{N'} = N'K'_0 + h'\sum_I S'_I + K'_1\sum_{I,J}S'_IS'_J + ...
\end{equation}
This can include cases where certain coefficients in the original system are zero but are not zero in the effective system. For example, in a spin system where every pair of spins interact, but there are no interactions of spin triples ($K_2\sum_{I,J,K}S_IS_JS_K, \hspace{.05in} K_2=0$), the effective system could have interactions of this type ($K'_2 \neq 0$).

Finally, condition (iii) ensures that there is a unique well-defined mapping of old degrees of freedom to the new ones. Furthermore, in the probabilistic case, this condition ensures it is well-defined by ensuring the sum of all the probabilities of transformed variables is 1. Furthermore, (iii) means that the partition function $Z_N$ is invariant under RG transformation \citep{Goldenfeld92}:
\begin{equation}
\begin{split}
    Z_N'[K'] = & Tr_{\{S'_I\}} \hspace{.05in}e^{\mathcal{H}'_{N'}\{[K'],S'_I\}} \\
    = & Tr_{\{S'_I\}} Tr_{\{S_i\}}  \hspace{.05in} P(S_i, S'_I)e^{\mathcal{H}_N\{[K],S_i\}} \\
    = & Tr_{\{S_i\}}  \hspace{.05in} e^{\mathcal{H}_N\{[K],S_i\}} \cdot 1 \\
    = & Z_N[K]
\end{split}
\end{equation}

As this is invariant, we can clearly see that the previously defined quantity $g$ for the free energy follows the relation:
\begin{equation}
\begin{split}
    g[K'] = & \frac{1}{N'} \log Z_{N'}[K'] \\
    = & \frac{1}{l^{-d}N} \log Z_{N}[K] \\
    = & l^d g[K]
\end{split}
\end{equation}

clearly conserving total energy as the free energy per degree of freedom has been scaled by the scaling factor $l^d$. One benefit of RG theory is that it is easier to establish approximate parameters [k'] as opposed to calculating the partition function.

%

\subsection{Fixed points and relevant parameters}

Now in order to analyse the behaviour of systems after repeated applications of the RG transformations, we consider the "flow" of parameters $[K^{(n)}]$. This group of constants from all sets of initial parameters $[K^{(0)}]$ is known as the renormalisation group flow. It has been noted \citep{Goldenfeld92} that the trajectories of parameters are usually attracted to certain fixed points, near to which the systems demonstrate characteristic scaling behaviour. 

Denote a fixed point of the RG transformation $R_l[K]$ in parameter space as $[K^*]$. It therefore has the property
\begin{equation}
    [K^*] = R_l[K^*]
\end{equation}

Looking at the correlation length at the fixed point, relation (\ref{correlation_length_relation}) gives
\begin{equation}
    \xi[K*] = \xi[K*]/l
\end{equation}
which implies $\xi$ is either $\infty$ of $0$. These two cases are denoted as "critical" and "trivial" fixed points respectively.

Within a vicinity of the fixed point, the parameters and Hamiltonian can be written
\begin{equation}
    K_n = K^*_n + \delta \hat{K}_n  ,   \hspace{.5in}    \mathcal{H} = \mathcal{H}^* + \delta\hat{\mathcal{H}}
\end{equation}
where $\delta$ is a small parameter.
Performing the RG transformation $R_l$ gives 
\begin{equation}
    R_l[K_n] = K'_n = K^*_n + \delta \hat{K}'_n
\end{equation}
where $\hat{K}'_n$ can be represented by the Taylor expansion:
\begin{equation}
    \hat{K}'_n = K^*_n + \sum_m \frac{\partial \hat{K}'_n}{\partial K_m}|_{K
=K^*} \cdot \delta K_m + O((\delta K)^2)
\end{equation}

so $\delta \hat{K}'_m = \sum_m M_{nm} \delta K_m$ where $M_{nm} =\frac{\partial \hat{K}'_n}{\partial K_m}|_{K=K^*}$ is the linearisation of the RG transformation near the fixed point. To analyse the RG flow in the vicinity of the fixed point $K^*$, the eigenvalues and eigenvectors of this linearisation are required. Denote the eigenvalues of $M^{(l)}$ (associated with transformation scale $l$) as $\lambda^{(l)}_i$ with corresponding eigenvectors $\textbf{v}^{(l)}_i$.
Due to the associativity of the RG transformation, we have the following property:
\begin{equation}
\begin{split}
    &M^{(l)}M^{(l')} = M^{(ll')} \\
    \implies & \lambda^{(l)}_i \lambda^{(l')}_i = \lambda^{(ll')}_i   
\end{split}
\end{equation}

Clearly we can see that $\lambda_i^1=1$ as $\lambda^{(l)}_i \lambda^{1}_i = \lambda^{(l)}_i$. Therefore, differentiating and setting $l'=1$, we get the differential equation:
\begin{equation}
    \begin{split}
    & l \frac{d\lambda^{(l)}}{dl} = y_i\lambda^{(l)} \\
    \rightarrow & \lambda^{(l)} = l^{y_i}
    \end{split}
\end{equation}
where $y_i$ is independent of $l$ and is to be determined.

The different values of $y_i$ for each eigenvector are important as they describe in what directions the components of $\delta K$ grow or shrink. 

\begin{enumerate}
    \item $y_i > 0$ ($|\lambda_i|>1$) implies $\delta K$ grows in the $\textbf{v}_i$ direction.
    \item $y_i = 0$ ($|\lambda_i|=1$) implies $\delta K$ is invariant in the $\textbf{v}_i$ direction.
    \item $y_i < 0$ ($|\lambda_i|<1$) implies $\delta K$ shrinks in the $\textbf{v}_i$ direction. 
\end{enumerate}
In case (1) they are called "relevant" eigenvalues/eigenvectors, in case (2), they are called "marginal" eigenvalues/eigenvectors, and case (3) are called "irrelevant" eigenvalues/eigenvectors. 

Therefore, the relevant and irrelevant directions determine the flow of parameters near a fixed point, and hence the critical behaviour. When the initial point is slightly away from the critical manifold of a fixed point (in parameter space), it will begin to travel towards the fixed point until it gets repelled according to the relevant eigenvectors.

This gives one part to the idea of universality as different initial conditions in parameter space can develop the same critical behaviour when $\mathcal{R}$ is repeatedly applied.

\subsection{Scaling of Variables}
Now that the critical behaviour has been explained using RG theory, it can be seen how RG leads to the scaling behaviour around criticality.

In the example of the Ising model, taking temperature and magnetism as $T$ and $H$, there are two relevant directions to consider: $t$ and $h$. As stated earlier, the singular part of the free energy density obeys the relation:
$$ f_s(t,h) = l^{-d} f_s(t',h')$$
Additionally, $T$ and $H$ both undergo individual course-graining processes $\mathcal{R}_l^T$ and $\mathcal{R}_l^H$.

Following the RG analysis, using the translated variables
\begin{equation}
    \begin{split}
        \Delta T =& T - T^* \\
        \Delta H =& H - H^*
    \end{split}
\end{equation}
we find the linearisation of the RG transformation 
$$\begin{pmatrix}
    \Delta T' \\
    \Delta H'
\end{pmatrix}
= M
\begin{pmatrix}
    \Delta T \\
    \Delta H
\end{pmatrix}$$
where
$$M = \begin{pmatrix}
    \frac{\partial R_l^T}{\partial T} &\frac{\partial R_l^T}{\partial H} \\
    \frac{\partial R_l^H}{\partial T} & \frac{\partial R_l^H}{\partial H}
\end{pmatrix}_{T=T^*, H=H^*}.
$$
Then writing the eigenvalues as before ($\lambda_l^t = l^{y_t}$, $\lambda_l^h = l^{y_h}$), the linearised RG transformation becomes
\begin{equation}
    \begin{pmatrix}
        t'\\
        h'
    \end{pmatrix}
    =
    \begin{pmatrix}
        \lambda_l^t & 0 \\
        0 & \lambda_l^h
    \end{pmatrix}
    \begin{pmatrix}
        t \\
        h
    \end{pmatrix}
\end{equation}

Now considering the free energy and correlation length, it can be seen that free energy evolves as
\begin{equation}
    f_s(t,h) = l^{-nd}f_s(t^{(n)},h^{(n)}) = l^{-nd}f_s(l^{ny_t}t, l^{ny_h}h)
\end{equation}
which is of the same form as in the block spin case (\ref{fs_eq2}).

Correlation length evolves according to (\ref{correlation_length_relation}) hence after n iterations of the renormalisation group transformation
\begin{equation}
    \xi(t,h) = l^n\xi(l^{ny_t}t, l^{ny_h}h)
\end{equation}

As the only restriction on $l$ is that $l>0$, taking $l^n = bt^{-1/y_t}$ for some arbitrary value b gives what is known as the static scaling hypothesis:
\begin{equation}
    \label{static_scaling_hyp}
    f_s(t,h) = t^{d/y_t}b^{-d}f_s(b, h/t^{y_h/y_t})
\end{equation}
The static scaling hypothesis is usually written in the form 
\begin{equation}
    \label{static_scaling_hypothesis_crit_exp}
    f_s(t,h) = t^{2-\alpha}F_{f}(\frac{h}{t^\Delta})
\end{equation}
where $F_f(x) = f_s(1,x)$, and $\alpha,\Delta$ are critical exponents \citep{Goldenfeld92}. Therefore, RG has provided a method to approximate the critical exponents from the eigenvalues of the linearised RG transformation:
\begin{equation}
    \begin{split}
        2-\alpha =& \frac{d}{y_t} \\
        \Delta =& \frac{y_h}{y_t}
    \end{split}
\end{equation}
\vspace{1cm}
\subsection{Universality Classes}

As described in the LTH, we are very interested in the idea of transferability across multiple different systems in order to improve training time. Therefore, the concept of universality classes is now briefly discussed in more detail. 

It has been seen that iterating the renormalisation group transformation causes the coupling constants to flow towards a fixed point, irrelevant of initial conditions. However, assuming the start point was off of the critical manifold, eventually, the irrelevant directions will have been "iterated out" \citep{UniversalitySulivan} leaving only the relevant directions controlling the macro-behaviour of the system. Therefore, different systems that share the same critical exponents will have the same relevant directions, and therefore demonstrate the same behaviour in this limit.

\vspace{1.5cm}
\section{DNNs and Iterative Magnitude Pruning (IMP)}
Now that the surrounding RG theory has been introduce, it must now be applied to the context of DNNs and pruning for winning lottery tickets. Therefore, the structures of neural networks and mechanisms of pruning are now set up.

The case that will be considered is that of a feed-forward DNN being pruned via iterative magnitude pruning (IMP). A feed-forward DNN consists of layers of neurons/units, with neurons in different layers being interconnected by weighted links. When an input is given to the first layer (input layer) of neurons, the following layers have values dependent on the values in the previous layer and the weighted connections. Denoting the weights connecting unit $i$ in layer $l$ and unit $j$ in layer $l+1$ as $\theta_{ij}$, the value of neuron $j$ in layer $n$ can be written $a_j = h(\sum_{i \in \textbf{n}} w_{ij}a_i + b)$, where $b$ is an added bias term and $h$ is a non-linear activation function (eg relu, tanh…).  

Once this process had continued through the network and reached the final/output layer, a final function can be applied to perform a task, such as data classification (sigmoid). In cases where supervised learning is used, the performance of the network can be measured via a loss function comparing the accuracy of the output compared to a known result (e.g \% of images correctly classified). However, in cases of unsupervised learning, the output accuracy can be quantified by comparing the performance to a desired result. This is controlled via a loss function $\mathcal{L}$. Gradient descent with respect to the loss is then performed on the weights in the network to minimise the loss. This stage is called backpropagation. Iterating this process (if hyperparameters are carefully selected) leads to improved performance at the task as the parameters are optimized to minimise the loss. 

As previously stated, training DNNs is a computationally expensive task which has motivated research into a variety of methods to increase efficiency of training. This includes recent work in the areas of transfer learning \citep{zhuang2020transfer_learning}, pruning \citep{vadera2021pruningmethods}, and utilising Koopman Operator Theory \citep{dogred20Koopman}.
We are most interested in reducing the over-parameterisation of DNNs and finding sparse subnetworks with similar performance to the original fully connected network and hence will be using pruning. IMP is one such method of doing this, removing weights with the smallest magnitude after t iterations of training the network. For more detailed analysis of the theory behind IMP, see \citep{elesedy2021lottery},\citep{maene2021understanding}.
An algorithm for IMP, removing one parameter from the network each iteration of pruning, can be written as follows \citep{elesedy2021lottery}:
\newpage

\textbf{Algorithm 1: IMP} \\
For loss function $\mathcal{L}:\mathbb{R}^p \rightarrow \mathbb{R}$, training time $T \in \mathbb{R}_+$, initial weights $w^{init} \in \mathbb{R}^p$ and $q<p$ iterations of pruning: 
\begin{itemize}
    \item Set $M = \mathbf{1}_P$
    \item For $k=0$ to $q$:\\
    Initialise $w^{(k)}(0) = Mw^{init}$ \\
    Train $\dot{w}^{(k)}(t) = -M \nabla L (w^{(k)}(t))$ for $t \in [0,T]$\\
    Set $i=$ argmin$_{j \in [p]} \{|w_j^{(k)}| : M_{jj} = 1\}$\\
    Set $M_{ii} = 0$
    \item Return $w^{(q)}(T)$
\end{itemize}

When applying the IMP algorithm to a DNN, it can either be applied across the whole network, or to individual layers, removing a specified proportion of non-zero weights each iteration. In practice, this is a computationally expensive process so to reduce the number of times that the network must be trained, a percentage of the remaining weights must be removed each iteration. This must be large enough to reduce the number of iterations, be small enough to not over-prune.
\\
\\
Work by Frankle and Carbin \citep{frankle2019lottery} shows the importance of resetting the weights to their original values before retraining. This is because when finding a potential lottery ticket, the initialisation is crucial to the efficiency of training the subnetwork and random initialisation results in decreased accuracy. This is because the mask is specific to the initialisation and therefore will only find a sparsified subnetwork with similar efficiency if the initial point is the same, as stated in the LTH.

Furthermore, once pruning is complete, the sparsified network must be fully trained again from the initial point in order to give accurate results. This is because the just pruned network is not fully optimised due to certain connections being set to zero, even if they were small.

The importance of the initialisation in terms of the LTH means that a winning ticket is comprised of both the mask $M$ and the weight initialisation $w(0)$.

\vspace{1.5cm}
\section{IMP as an RG}

\subsection{Proof IMP is an RG scheme}

In order to use the RG tools defined in the previous section for the purposes of analysing IMP, it must be shown that IMP can be seen as a renormalization group transformation. To do this, comparisons are made between DNNs with IMP and the 2D Ising model in order to motivate the construction of a similar course graining process to that of the block spins. It can in fact be shown that they are both systems of an equivalent form. Although some features of RG theory are not yet fully understood, such as correlation length, following the work of Redman et.al.\citep{redman}, we can prove IMP is a renormalisation group operator and hence gather useful tools from this link.
\\
\\
On a conceptual basis, it is clear to see how a DNN under IMP can be visualised as an Ising model with course-graining. 
A DNN can be visualised as a lattice structure of nodes with each node characterised by the activation value $a_j = h(\sum_{i \in \textbf{n}} w_{ij}a_i + b)$. Then each node is connected by a weighted connection to the nodes in the neighbouring layer. 
\\
\\
Taking a more formal approach, we can immediately see the similarities between the effects of IMP $\mathcal{I}$ operator and the $\mathcal{R}$ operator from the block spin system. Consider a DNN with parameters $\pmb{\theta}$, activations $\textbf{a}$, and loss function $\mathcal{L}$. Applying the IMP operator $\mathcal{I}$ gives:
\begin{equation}
    \mathcal{I}\mathcal{L}(\textbf{a}, \pmb{\theta}) = \mathcal{L}(\textbf{a}', \pmb{\theta}') = \mathcal{L}(\textbf{a}', \mathcal{T}\pmb{\theta})
\end{equation}
This is analogous to the $\mathcal{R}$ operator acting on the Hamiltonian in the classical spin system, taking $\mathcal{L}$ as analogous to $\mathcal{H}$, $\textbf{a}$ for $\textbf{s}$ and $\pmb{\theta}$ for $\textbf{K}$. Here the operator $\mathcal{T}: \mathbb{R}^N \rightarrow \mathbb{R}^N$ (N being the total number of parameters in the network) is a composition of a masking operator $\mathcal{M}$ and a training operator $\mathcal{F}$, thus only retraining the non-pruned parameters. $\mathcal{M}$ is defined by the chosen method of pruning, which in our case is magnitude pruning, and $\mathcal{F}$ is defined by the optimizer used and whether the weights are being rewound to their initialisation. 
\\
\\
Hence the full process of IMP can be written:

\begin{equation}
    \mathcal{I}^n\mathcal{L}(\textbf{a}^{0}, \pmb{\theta}^{0}) = \mathcal{L}(\textbf{a}'^{n-1}, \pmb{\theta}'^{n-1}) = \mathcal{L}(\textbf{a}'^{n-1}, \mathcal{T}^{n}\pmb{\theta}^{0})
\end{equation}

This sequence of evolving parameters $\{\pmb{\theta}^{i}\}_{i}$, determined by the eigenvectors of $\mathcal{T}$, gives an equivalent to the RG flow described before. In the context of IMP, we call this sequence the IMP flow. 
\\
\\
In order to show that this is an RG scheme, we consider analogous elements of the system to that of the block spin model. Analogous to the spins $\textbf{s}$ is the activations $\textbf{a}$ of each unit in the model; the couplings between spins are analogous to the model parameters $\pmb{\theta}$; and the Hamiltonian $\mathcal{H}$ is analogous to the loss function $\mathcal{L}(\textbf{a},\pmb{\theta})$ \citep{redman}.
\\
\\
For a general DNN, we define the activation $a_j^{(i)}$ of the j-th unit in layer i as:
\begin{equation}
    a_j^{(i)} = h[ \sum_k g_k(\textbf{a}, \pmb{\theta}) ]
\end{equation}
where $g_k(\textbf{a}, \pmb{\theta})$ are the functions that determine the effect of other parameters and activations on $a_j^{(i)}$ and $h$ is the selected activation function. For example, in a feed-forward DNN, each activation is the sum of the product of the previous layers' activations and the weights of parameters connecting them. There is also an additional bias term added to each activation. Thus, there are two functions:
\begin{enumerate}
    \item Bias: $g_0 = \theta_j^{(i)}$
    \item Weighted input from previous layer: $g_1  =\sum_{k=1}^{N^{i-1}} \theta_{jk}^{(i)} a_k^{(i-1)}$
\end{enumerate}
Where $\theta_j^{(i)}$ are biases for unit $j$ in layer $i$, $\theta_{jk}^{(i)}$ are weights connecting $a_k^{(i-1)}$ to $a_j^{(i)}$ and $N^{i-1}$ is the number of activations in layer ($i$-1).
\\
\\
Therefore after applying the IMP operator $\mathcal{I}$, the parameters $\pmb{\theta}$ have been transformed hence giving:
\begin{equation}
    a_j^{\prime(i)} = h[ \sum_k g_k(\textbf{a}', \mathcal{T}\pmb{\theta}) ]
    = h[ \sum_k g_k(\textbf{a}', \mathcal{F} \circ \mathcal{M}\pmb{\theta}) ]
\end{equation}

Therefore, we can define the associated projection operator for $\mathcal{I}$ in the same way as we did for the block spin case.

\begin{equation}
    P(a_j^{(i)}, a_j^{\prime(i)}) = \prod_I \delta(a_j^{\prime(i)} - h[\sum_k g_k(\textbf{a}',\pmb{\theta}')])
\end{equation}

To ensure this is an RG projection operator, it can verified that it meets the previously defined requirements of a projection operator (\ref{conditions}): \citep{redman}

\begin{enumerate}[i]
    \item As $P()$ is a product of Kronecker deltas, it clearly satisfies (i) $P(s_i, s_I') \geq 0$
    \item Property (ii) is met as the process of IMP only removes parameters from the model. Therefore, until the point where every weight in a layer has been pruned (layer collapse), the activations $a_j^{(i)}$ still have the same form, and thus the loss function is still of the same form.
    \item Property (iii) is met when the masking and refining operators $\mathcal{M}$ and $\mathcal{F}$ are deterministic. This requires fixing the seed so that the sampling order of test and train data (in the case of supervised learning) is fixed for every epoch. This ensures that $P$ is a unique projection operator.
\end{enumerate}

As the constructed projection operator fulfils the requirements of a renormalisation group projection operator, it has been shown that $\mathcal{I}$ is and RG operator.

\subsection{IMP Flow}

Now that working in an RG framework has been justified, we can use the tools available in this theory to analyse IMP behaviour. Clearly understanding the critical manifold is of great importance as it can give a clear indication of which parameters are relevant and irrelevant during pruning. The parameters $\theta_i$ are transformed by the transformation $\mathcal{T}$. Therefore, by finding eigenfunctions of  $\mathcal{T}$, we can find the relevant and irrelevant eigenvalues/eigenvectors \citep{redman}. This is important in the context of the LTH as different models which have the same relevant eigenvectors $\textbf{v}_i$ with $\lambda_i>1$, should share the same subset of parameters which remain after pruning.

One proposed eigenfunction of $\mathcal{T}$ by \citep{redman} which will be considered relates to the proportion of total magnitude of parameters remaining in each layer after n iterations of IMP.
Consider the function 
\begin{equation}
\label{eigenfunction}
    M_i(n) = \frac{\sum_{j}|\theta^{(i)}_j(n)|}{\sum_i\sum_j|\theta_j^{(i)}(n)|}
\end{equation}
where $\theta_j^{(i)}(n)$ is the jth weight in layer i after n iterations of IMP.
In order to find the eigenvalues associated with this eigenfunction, we use the following relation 
\begin{equation}
    M_i(n+1) = \lambda_i M_i(n)
\end{equation}
Therefore, to find $\lambda_i$ we invert this to get $\lambda_i = \frac{M_i(n+1)}{M_i(n)}$. As in the process of defining renormalisation groups, in order to having a meaningful value for the relevance of each layer, we write $\lambda_i=l^\sigma_i$ where $\sigma_i$ is invariant of the course-graining $l$. Therefore, $\sigma_i$ may be used to compare behaviour with different models which may use a different pruning rate $l$.

\vspace{1.5cm}
\section{Neural Network DE Solvers}
To support the use of RG tools in finding universal lottery tickets, we apply this theory to the case of neural networks for solving differential equations. This is an important field as, unlike traditional solvers which generate a solution at a range of individual points, DNNs can provide a differentiable solution in a closed analytic form \citep{Lagaris_1998_NNDEs}. Another advantage is that they are very capable when solving high dimensional problems while exploiting parallel computing. They have been found to be applicable to a wide range of PDEs and dynamical systems \citep{Han_2018},\citep{Mattheakis_2022},\citep{Lagaris_1998_NNDEs},\citep{Lu_2021_deep-o-net},\citep{SIRIGNANO20181339DGM},\citep{Mattey_2022_PINNS},\citep{hornung2020spacetime} so RG theory can allow us to find "similar" systems amongst these to help improve training times by utilising transferable lottery tickets.

A useful example to consider is Hamiltonian Neural Networks (HNNs). These are designed to use unsupervised learning to solve problems while conserving physical properties - such as energy - and are applicable to wide range of problems of different complexities in Hamiltonian mechanics \citep{TaylorMechanics}. Even though these networks are less complex than some solvers, they provide a useful insight into how RG tools can be used to analyse IMP and universality in scientific solvers for well-posed problems.

As an example, we perform IMP on a existing neural network. The codebase for this can be found here \citep{eddie_git}. Work by Mattheakis et.al.\citep{Mattheakis_2022} explores the behaviour of Hamiltonian neural networks being used to solve Hamilton's equations in order to obtain equations of motion for two different dynamical systems - a non-linear oscillator and a H\'enon-Heiles system. For more analysis of these models see \citep{Mattheakis_2022, dog23_3, dog23_2}. 

\subsection{Nonlinear Oscillator}
The first case considered is a neural network designed to solve the nonlinear one-dimensional an-harmonic oscillator described by the Hamiltonian
\begin{equation}
    \mathcal{H} = \frac{p^2}{2} + \frac{x^2}{2} + \frac{x^4}{4}
\end{equation}
where mass and natural frequency are taken to be 1. The Hamiltonian $\mathcal{H}$ represents the total energy $E$ of the system while the associated Hamilton's equations for this system, given by $\dot{x} = \frac{\partial H}{\partial p}$ and $\dot{p} = -\frac{\partial H}{\partial x}$, are:
\begin{equation}
    \dot{x} = p, \hspace{1in} \dot{p} = -(x+x^3)
\end{equation}

These are solved by the neural network by minimising the loss function

\begin{equation}
    L = \frac{1}{K} \sum_{n=1}^K[(\dot{\hat{x}}^{(n)} - \dot{\hat{p}}^{(n)})^2 + (\dot{\hat{p}}^{(n)} + \dot{\hat{x}}^{(n)} + (\dot{\hat{x}}^{(n)})^3)^2]
\end{equation}
which is the mean square error from Hamilton's equations.
\\
\\
The neural network used to solve this optimisation problem has 2 hidden layers, each with 50 neurons, and an output layer of 2 neurons (corresponding to the two degrees of freedom in the problem). Therefore, there are a total of (50 + 50x50 + 50x2 =) 2650 weights in the neural network and 102 biases. The hyperparameters were set to the same values as used by Mattheakis et.al. (lr=$8 \cdot 10^{-3}$, $5 \cdot 10^4$ epochs, ...) as these were shown to achieve a high accuracy solution which we aim to maintain during pruning.
\\
\\
For now, we will ignore the effect of pruning biases directly, as much of that behaviour will be implicitly included when pruning weights. This is because when weights connecting to a neuron are pruned, the bias at that neuron becomes less significant. 
\\
\\
In order to obtain an effective theory of IMP in neural networks for solving DEs, different pruning procedures are performed in order to understand different scalings that exist in the model. First, the effect of pruning individual layers is explored. Therefore, to carry out the previously explained algorithm in this case, we only use the masking operator over parameters in the desired layer - by which the process for one iteration of pruning on layer $l$ becomes:
\begin{itemize}
    \item Fully train neural network
    \item Prune $p\%$ of remaining weights in layer $l$
    \item Reset remaining weights to initial values
    \item Fully train remaining parameters
\end{itemize}

To ensure IMP is implemented correctly, the weights are reset to the initial random values after each pruning iteration. Furthermore, to ensure that the pruned weights are no longer trained, the corresponding gradient is set to zero during the backpropagation stage.
\\
\\
We expect to find similar behaviour to that found by Rosenfeld et.al.\citep{https://doi.org/10.48550/arxiv.2006.10621}, namely that error scales with a power law relationship $\epsilon = cd^{-\gamma}$ within in a critical region of pruning. This would support the use of RG theory to analyse IMP. For each layer, numerical experiments have been performed, pruning at 1\%, 5\% and 10\% each iteration to gain an understanding of how pruning rate can affect the efficiency of IMP. Work by Vandersmissen et.al.\citep{vandersmissen2023layerLTH} has suggested that the lottery ticket hypothesis is not affected by pruning rate so we expect to see similar behaviour but at different resolutions. Additionally, during each experiment, the layer was only pruned until 10\% of weights remain to prevent full layer collapse. Below are the results for these experiments:

\begin{figure}[H]
\centering
\begin{subfigure}[b]{.55\linewidth}
    \centering
    \includegraphics[width=\linewidth]{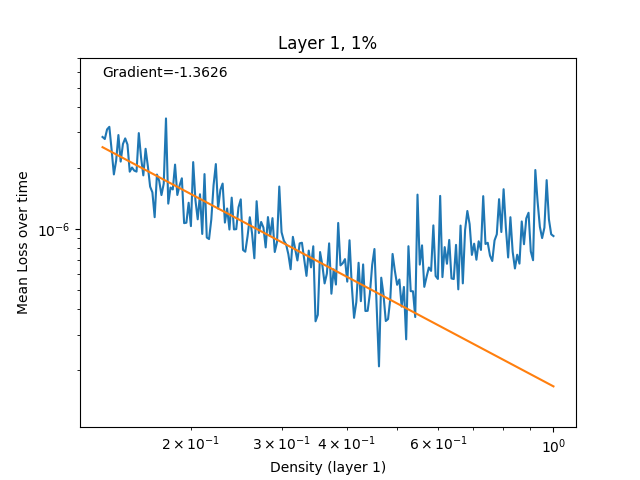}
    \caption{Layer 1, 1\%}
    \label{layer1_1}
\end{subfigure}
\begin{subfigure}[b]{.55\linewidth}
    \centering
    \includegraphics[width=\linewidth]{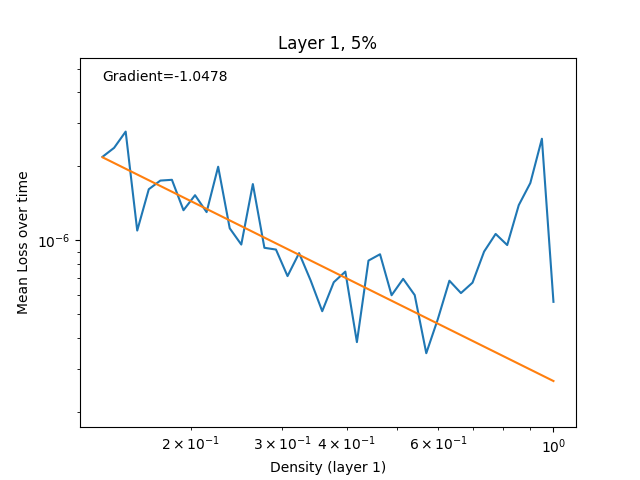}
    \caption{layer 1, 5\%}
    \label{layer1_5}
\end{subfigure}
\begin{subfigure}[b]{.55\linewidth}
    \centering
    \includegraphics[width=\linewidth]{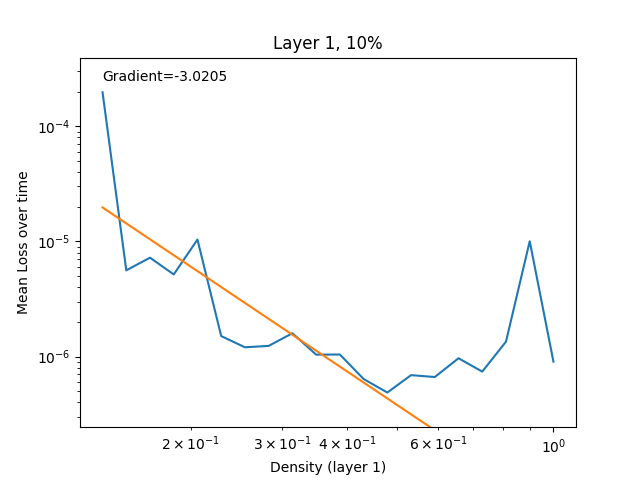}
    \caption{layer 1, 10\%}
    \label{layer1_10}
\end{subfigure}
\caption{Pruning of layer 1 with an initial 50 weights}
\end{figure}

\begin{figure}[H]
\centering
\begin{subfigure}[b]{.55\linewidth}
    \centering
    \includegraphics[width=\linewidth]{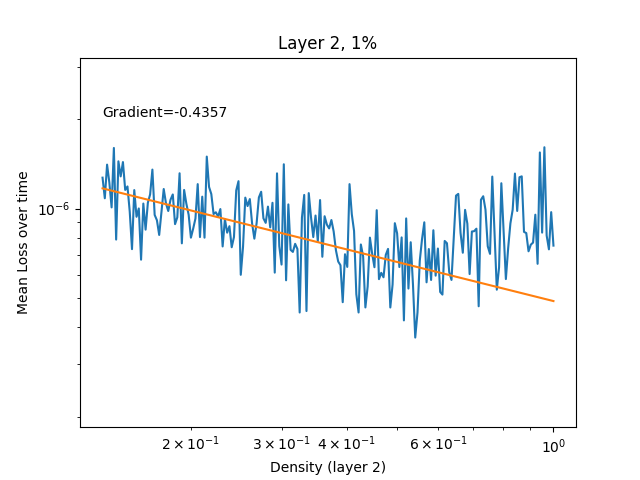}
    \caption{layer 2, 1\%}
    \label{layer2_1}
\end{subfigure}
\begin{subfigure}[b]{.55\linewidth}
    \centering
    \includegraphics[width=\linewidth]{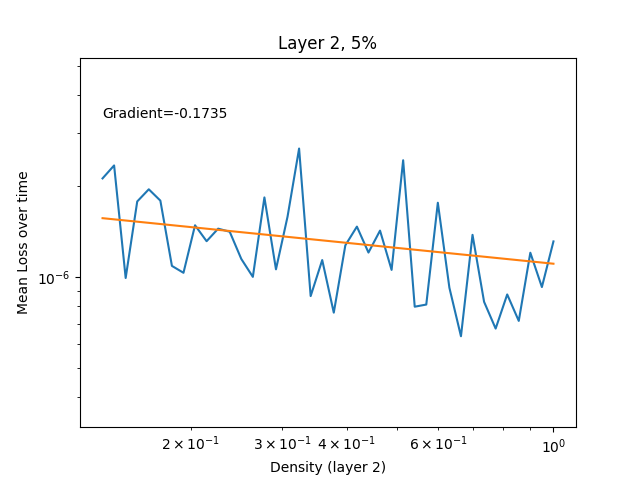}
    \caption{layer 2, 5\%}
    \label{layer2_5}
\end{subfigure}
\begin{subfigure}[b]{.55\linewidth}
    \centering
    \includegraphics[width=\linewidth]{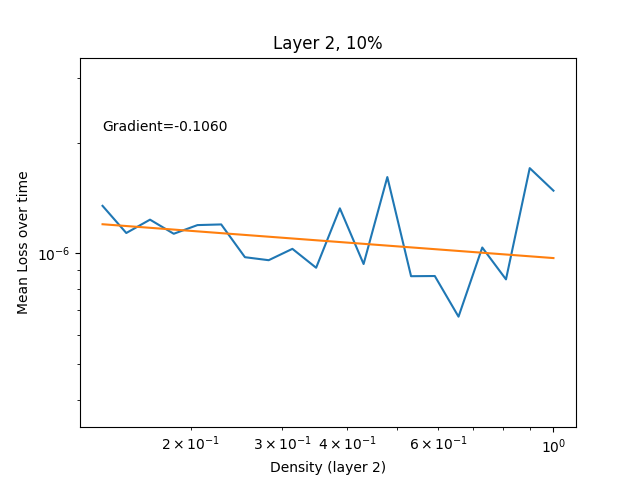}
    \caption{layer 2, 10\%}
    \label{layer2_10}
\end{subfigure}
\caption{Pruning of layer 2 with an initial 2500 weights}
\end{figure}

\begin{figure}[H]
\centering
\begin{subfigure}[b]{.55\linewidth}
    \centering
    \includegraphics[width=\linewidth]{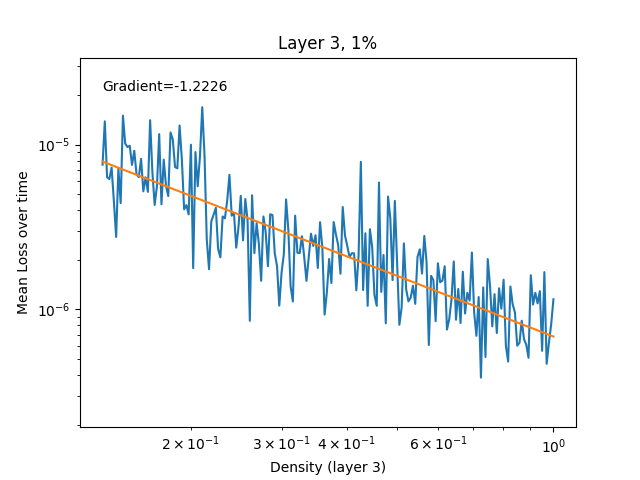}
    \caption{layer 3, 1\%}
    \label{layer3_1}
\end{subfigure}
\begin{subfigure}[b]{.55\linewidth}
    \centering
    \includegraphics[width=\linewidth]{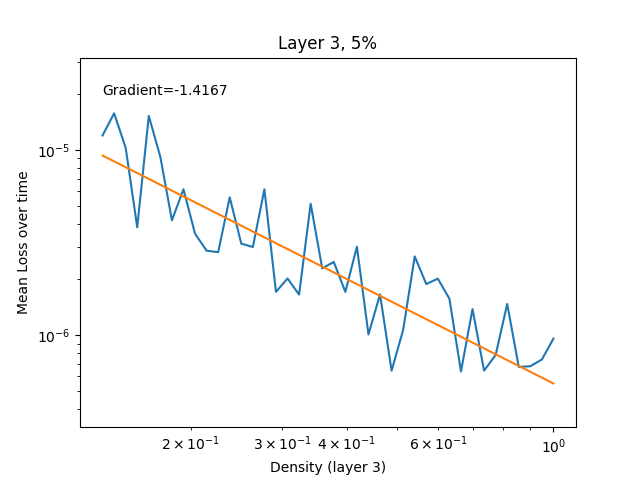}
    \caption{layer 3, 5\%}
    \label{layer3_5}
\end{subfigure}
\begin{subfigure}[b]{.55\linewidth}
    \centering
    \includegraphics[width=\linewidth]{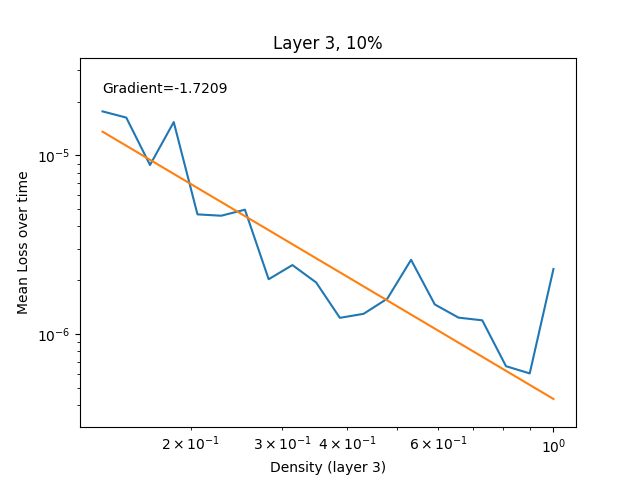}
    \caption{layer 3, 10\%}
    \label{layer3_10}
\end{subfigure}
\caption{Pruning of layer 3 with an initial 100 weights}
\end{figure}

Here we can clearly begin to see the expected behaviour in loss. As more of the parameters are pruned, we see that the network is less able to learn how to solve the equations accurately. From the finest level of pruning (1\%) we can see how there is a critical percentage of weights that can be pruned in each layer before we see the power-law relationship between layer density and the error in the solution (e.g layer 1 $\approx$ 50\%).
This exemplifies the LTH as there are clearly redundant weights which can be removed without negatively impacting on accuracy. This is because other weights in the network are able to account for the lack of weights in other layers. This explains why, when pruning layers individually, the deeper layers can withstand less pruning before an increase in error. When weights are removed front the top of the network, there are many layers below which can account for this. However, when weights are removed from deeper in the network, information is being lost from layers above which cannot be accounted for below.

Using the gradient of the graphs in the interval where the power-law relationship is visible, the critical exponents corresponding to error when pruning each layer individually can be observed. The case of 1\% pruning is used as this gives the finest resolution of the behaviour in each layer.

\begin{table}[H]
    \centering
    \begin{tabular}{|c|c|} \hline
      layer     & $\gamma$ \\ \hline
      layer 1   & 1.36 \\ \hline
      layer 2   & 0.44 \\ \hline
      layer 3   & 1.22 \\ \hline
    \end{tabular}
    \caption{Critical exponents retaining to error as a function of density of on individual layer.}
    \label{tab:my_label}
\end{table}

This gives an idea of each layer's sensitivity to pruning. For example, pruning either the input or output layer has the most significant impact on the accuracy of the model.

Furthermore, comparing the different pruning rate, we see that each rate shows the same behaviour but to a different resolution. Therefore IMP should be able to find a winning ticket for a range of pruning rates, so long as it is not too high. This corroborates findings by Vandermissen et.al.\citep{vandersmissen2023layerLTH}.

So far, only the "macro" effect of pruning on the error from the network one layer at a time. 
The next experiment to be considered is that where all layers are pruned together. This gives a more detailed picture of how the coupling in the network develops during pruning, and therefore relates strongly with the RG theory previously explained. Therefore, comparing critical exponents from this experiment will be able to be used to give some idea of transferability.
For this experiment, at each pruning stage, the 5\% of smallest weights from the entire network are set to 0. This is the most useful case in finding potential lottery tickets as one aims to remove as many parameters from the network as possible without negatively affecting accuracy. Similarly to in the previous experiments, each layer is no longer pruned after reaching 5\% density to prevent layer collapse.

As described early in section 4, this scenario directly compares to a 2D Ising model where we considered a lattice of activations coupled by weights in the network. 
The results we find for this case give a fuller picture of the potential of IMP in finding winning tickets and the power-law scaling of error during pruning.

\begin{figure}[H]
    \centering \includegraphics[width=0.7\linewidth]{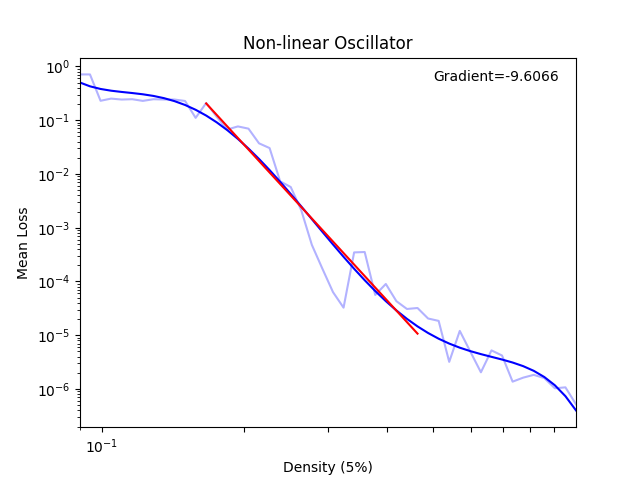}
    \caption{Error of NN solving the non-linear oscillator system during IMP pruning at 5\%.}
    \label{fig:my_label}
\end{figure}

In these results, the behaviour of power law scaling within a certain critical region can clearly be observed, which helped motivate the use of RG theory to describe IMP. 
Furthermore, this corroborates findings from Rosenfeld et al.(2021) \citep{pmlr-v139-rosenfeld21a} which show that different machine learning models have three different periods during pruning. The first of these is the low-error plateau where error is similar to that of the fully-connected network. The end of this region is the point of interest when trying to find a winning lottery ticket. The next stage is where the power-law scaling is observed. As stated earlier, this is modelled by $\epsilon \approx cd^{-\gamma}$. The final segment is where the error levels off at a high value as the network is no longer to learn anything significant. This is called the high-error plateau $\epsilon^\uparrow$. \citep{pmlr-v139-rosenfeld21a}

Extracting the gradient of the log-log graph for the power-law scaling section of the pruning process gives the critical exponent for the error as a function of density. So, for this system, the critical component $\gamma=9.61$.

\subsection{IMP Flow}
For the case where parameters were pruned from any layer in the model, we use the procedure previously defined in (\ref{eigenfunction}) to obtain a value for $\sigma_i$ in each layer of the DNN to gather a representation of the significance of each layer in the pruning process. 

For the Non-linear Oscillator, the $\sigma_i$ from each eigenvalue for each hidden layer are as follows. 
\begin{table}[H]
    \centering
    \begin{tabular}{|c|c|}
    \hline
    $\sigma_1$ & 0.313 \\ \hline
    $\sigma_2$ & -0.010 \\ \hline
    $\sigma_3$ & 0.736 \\ \hline
\end{tabular}
    \caption{Critical exponents corresponding to magnitude remaining in each layer of the non-linear oscillator system throughout pruning.}
    \label{tab:my_label}
\end{table}

These results were collected from pruning 5\% of the networks weights during each iteration of IMP, however they are independent of the pruning rate. Clearly we see that this supports the data from the previous section that layers 1 and 3 are relevant whereas layer 2 is not relevant (using the definition of relevant and irrelevant eigenvalues from previously).

\subsection{Chaotic H\'enon-Heiles}
As previously stated, one of the most useful links between the LTH and RG theory is the concept of universality. In application, once one successful mask has been found to give efficient results on one network, it may be transferred to another network without having to re-perform the computationally expensive pruning process.

Work by Morcos et al.\citep{morcos2019ticket} has already shown that for certain tasks such as computer vision, winning tickets can be transferred across different tasks/models. Furthermore, work by Redman et.al.\citep{redman} has suggested that in the cases where this is possible, the blocks/layers have similar relevance ($\sigma_i$) and can therefore be considered in the same universality class.  

Therefore, in an attempt to find another model in the same universality class as the non-linear oscillator solver analysed, we analyse a new neural network DE solver for a H\'enon-Heiles oscillator system. \citep{Henon-Heiles}

The H\'enon-Heiles system is a chaotic system describing the non-linear planar trajectory of body around a galactic centre. The degrees of freedom in the system are $(x,y,p_x,p_y)$, representing position and momentum respectively.

The total energy of the system is given by the following Hamiltonian \citep{Mattheakis_2022}:

\begin{equation}
    \mathcal{H} = \frac{1}{2}(p_x^2+p_y^2) + \frac{1}{2}(x^2+y^2) + (x^2y-\frac{1}{3}y^3)
\end{equation}

Therefore we can extract Hamilton's equations:

\begin{equation}
\begin{split}
    \dot{x} = \frac{\partial\mathcal{H}}{\partial p_x} &= p_x \\
    \dot{y} = \frac{\partial\mathcal{H}}{\partial p_y} &= p_y \\
    \dot{p_x} = -\frac{\partial\mathcal{H}}{\partial x} &= -(x+2xy) \\
    \dot{p_y} = -\frac{\partial\mathcal{H}}{\partial y} &= -(y+x^2-y^2)
\end{split}
\end{equation}

As in the one-dimensional case, the loss function is taken to be the mean squared error from Hamilton's equations in order to conserve the Hamiltonian:

\begin{equation}
    L = \frac{1}{K} \sum_{n=0}^K [ (\dot{\hat{x}}^{(n)} - \dot{\hat{p}}_x^{(n)})^2 + (\dot{\hat{y}}^{(n)} - \dot{\hat{p}}_y^{(n)})^2 + (\dot{\hat{p}}_x^{(n)}+\hat{x}^{(n)} + 2\hat{x}^{(n)}\hat{y}^{(n)})^2 + (\dot{\hat{p}}_y^{(n)}+\hat{y}^{(n)}+(\hat{x}^{(n)})^2 - (\hat{y}^{(n)})^2)^2 ]
\end{equation}

The neural network used to solve this problem is of a very similar structure to that of the non-linear oscillator. However, it has some slight differences. As before, two hidden layers of 50 neurons are used, but the output layer is now 4 nodes instead of 2, to account for the extra dimension in the problem. In order to keep the rest of the system similar, the same activation function ($\sin(\cdot)$) and optimiser (Adam) are used. With both structural similarities and similarities in the problems being solved, it is hoped that the two systems considered are in the same universality class. 

Repeating the procedure as for the non-linear oscillator system, the IMP flow of parameters in the new system is found. The $\sigma_i$ corresponding to each eigenvalue is found to be as follows:

\begin{table}[H]
    \centering
    \begin{tabular}{|c|c|c|}
        \hline
         & Non-linear Oscillator & H\'enon-Heiles Oscillator \\ \hline
        $\sigma_1$ & 0.313 & 0.375\\ \hline
        $\sigma_2$ & -0.010 & -0.012 \\ \hline
        $\sigma_3$ & 0.736 & 0.630 \\ \hline
    \end{tabular}
    \caption{Critical exponents corresponding to magnitude remaining in each layer of both systems.}
    \label{tab:compare e-vals}
\end{table}

Clearly we can see that the two systems have the same relevant layers and similar values for $\sigma_i$ in each layer. This therefore suggests that a winning ticket can be transferred between the two models. 

Plotting the error as the model is pruned, we see that it follows a similar behaviour to that of the non-linear oscillator in that the periods of low-error plateau, power-law scaling, and high-error plateau fall within similar regions of density.

\begin{figure}[H]
    \centering
    \includegraphics[width=.8\linewidth]{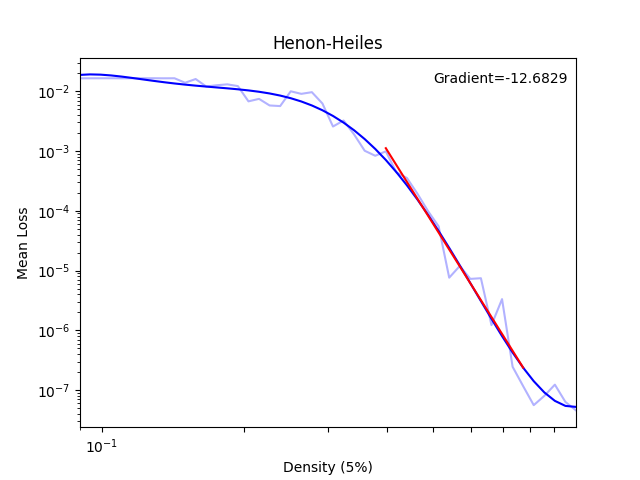}
    \caption{Error of NN solving H\'enon-Heiles system during IMP at 5\% pruning rate}
    \label{fig:my_label}
\end{figure}

Furthermore, relating IMP to pruning, the two systems have similar values of critical exponent $\gamma$, from equation (\ref{crit_exp}), for the macro-observable error. Therefore, according to RG theory, the two systems should exhibit the same behaviour under the IMP operator.

\subsection{Universality}
So far, it has been observed that these two similar systems have the same relevant layers and similar $\sigma_i$ and $\gamma$ values. This implies that a winning ticket in one case can be transferred between the two systems and still provide improved performance. However, the difference in architecture means that the mask and initialisation cannot simply be carried between the two networks. Therefore, we utilise a strategy as suggested by experiments around the Elastic Lottery Hypothesis (ELH) \citep{chen2021elastic}.

The Elastic Lottery Hypothesis is an extension of the LTH as it suggests that a winning ticket for a certain system can be transferred to a different system with different architecture while still maintaining performance and improving training time. In order to do this, winning tickets must be "stretched" or "squeezed" into the new architecture. For the case being considered here, the winning ticket for the non-linear oscillator must be stretched into the wider H\'enon-Heile architecture. The suggested method for doing this, which has shown success in other types of neural network \citep{chen2021elastic}, is by duplicating a block/layer of the winning ticket to extend it. Therefore, a winning ticket for the non-linear oscillator must have the third layer duplicated in order to be applied to the H\'enon-Heiles system.

Now using this method to stretch the winning ticket, the masks found while pruning the non-linear oscillator are now applied to the H\'enon-Heiles network. Due to the nature of two different systems, the effect of certain hyperparameters must also be investigated. Although hyperparameters such as width, depth and learning rate are easily kept the same, the time span for which the system is being solved over is also important. In the non-linear oscillator case, the network is solving Hamilton's Equations for 200 equally spaced points on the time interval $[0,4\pi]$. Therefore, it would be useful to know whether the lottery ticket for this model is applicable to solutions over multiple time intervals.  Plotting the results below, we fit a curve to gather the general trend for each system.

\begin{figure}[H]
    \centering
    \includegraphics[width=.8\linewidth]{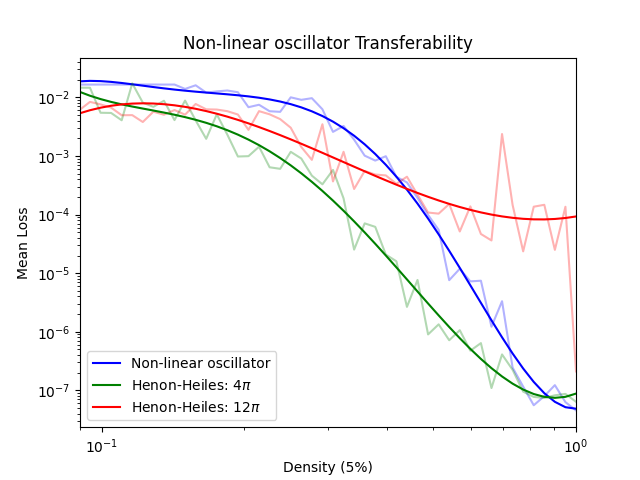}
    \caption{Transferability of winning ticket to H\'enon-Heiles system solving over different integration times.}
    \label{fig:my_label}
\end{figure}

What can be seen is that a lottery ticket is most effective when applied to a H\'enon-Heiles system solving over a similar time period. This is because the complexity and hyperparameters of the original system must have the capability of solving the new system effectively. Solving over a longer integration time requires a network of greater complexity and size. Hence the periods of low-error plateau, power-law scaling, and high-error plateau are much less visible. The optimal integration time to be used is dependent on the system being considered. The best candidate would be one which provides a similar error for the fully-connected network.

Comparing the results for $t=4\pi$ with existing experiments in the field of computer vision, it is seen that transferability is not as strong in neural network DE solvers. One potential reason for this is that the neural networks in this case are far smaller than in convolutional networks. For example, Resnet-50 models contain several millions of trainable parameters, whereas simple DE solvers contain less than 10,000. Therefore, there is a more clearly defined region for the low-error plateau as each trainable parameter is far less significant in the entire network, and hence has less effect when removed.

Repeating this process in the other direction, we take masks throughout the pruning process of the H\'enon-Heiles network and apply them to the non-linear oscillator system. Due to the differing architectures of the systems, the ticket must now be "squeezed" as opposed to stretched into the new network. This involves removing blocks from the lottery ticket. In this case, half of the final layer of weights will be removed so that it outputs 2 values as opposed to 4. Work by Chen et.al.\citep{chen2021elastic} suggests that systems are not sensitive to whether consecutive or non-consecutive blocks are removed. Therefore we decided to remove the second and fourth blocks, as these correspond to position and momentum in the second dimension. 

\begin{figure}[H]
    \centering
    \includegraphics[width=0.8\linewidth]{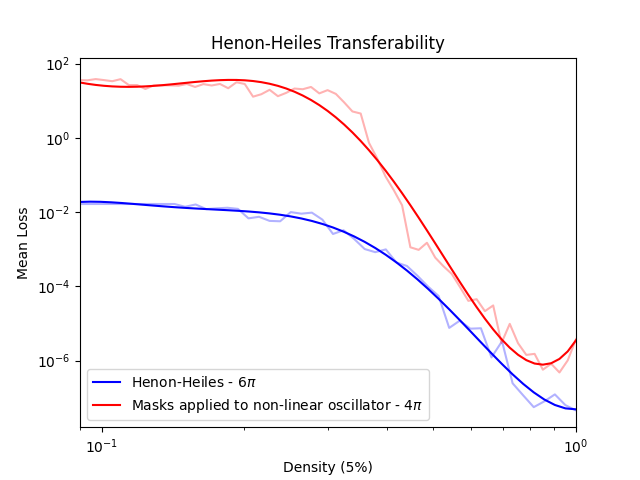}
    \caption{Transferability of winning ticket to non-linear oscillator system}
    \label{fig:transfer2}
\end{figure}

As we can see, the transferability works in both directions with systems showing similar behaviour at similar densities. Again, the value of error is dependent on the integration time. As transferability works in both directions, it is strong evidence that the two systems are in fact in the same universality class. 

\vspace{1.5cm}
\section{Conclusions}
By comparing DNNs under IMP with extensively studied thermodynamic systems, we have been able to motivate the use of RG theory to explain neural network behaviour during pruning. Following the work of Redman et.al.\citep{redman}, we were able to formally show that IMP is a renormalisation group transformation. As renormalisation group theory has been the primary method of explaining universality across different systems sharing the same critical exponents during a phase transition, we aimed to extend existing results in the field of computer vision to neural network DE solvers. In the context of the winning lottery ticket hypothesis, this involved transferring lottery tickets from a Hamiltonian Neural Network solving Hamilton's equations in a non-linear oscillator system to one solving Hamilton's equations in the H\'enon-Heiles system. 
Firstly, by pruning individual layers of the non-linear oscillator system, we saw that the system is in fact over-parameterised and therefore a winning ticket does exist to increase training time without negatively impacting accuracy.  Then pruning both the non-linear oscillator and the H\'enon-Heiles system showed that they both have the save relevant layers (shown by $\sigma_i$ values). Hence work by Redman et.al. suggests that these two systems will show transferability of winning lottery tickets. Experiments for this showed that transferring lottery tickets preserves the pruning behaviour in the new system. However, the success of this depends on the integration time used in each system. As winning lottery tickets only enable a neural network to achieve the accuracy of the fully connected network, the fully connected network must be capable of reaching a high enough accuracy. Therefore, transferring winning lottery tickets is most successful when the new system is similarly over-parameterised and has a similar accuracy to the original system. To solve for larger integration times, larger networks with more complex architectures must be used.  The issue when doing this is that either the winning ticket must be stretched to fit the new architecture, or the system must be pruned again to find a new winning ticket. The experiments performed provide further supporting evidence for results found by Chen et.al.\citep{chen2021elastic} on the elastic lottery hypothesis that duplicating winning tickets into wider network architectures can provide high accuracy during transfer.

\subsection*{Future Directions}
There are many steps that can be taken in furthering the knowledge of transferability of winning lottery tickets between different neural network models to solve DEs. Firstly, the behaviour of deeper networks and whether this leads to more disparities in layer relevance. Currently, we have only looked at simple networks with 2 hidden layers, but more complex systems and longer integration times will require more complex architectures, in which pruning will be even more important to reduce computation time. This also enables research into transferability between models of different depths, extending knowledge around the Elastic Lottery Hypothesis.
Additionally, considering more systems will allow for them to be categorised by their critical exponent $\gamma$ to find different classes of "similar" solvers.
As using RG theory to analyse pruning IMP has shown promising results, filling in gaps in the connection between the two, such as correlation functions, could give new insights into the pruning process.

\bibliographystyle{unsrtnat}

\bibliography{refs}
\addcontentsline{toc}{section}{Bibliography}

\end{document}